\documentclass[11pt,onecolumn]{IEEEtran}

\usepackage{amsmath}
\usepackage{graphicx}
\usepackage{color}
\usepackage{algorithm}
\usepackage{algpseudocode}
\usepackage{multirow}
\usepackage{tikz}
\usepackage{verbatim}
\usepackage[margin=0.5in]{geometry}
\usepackage{textcomp}
\usepackage{pgfplots}
\usepackage{mathrsfs}

\definecolor{darkred}{rgb}{0.7,0,0}
\definecolor{darkblue}{rgb}{0,0,0.5}
\definecolor{darkgreen}{rgb}{0,0.7,0}

\begin{document}

\title{Voting Network for Contour Levee Farmland Segmentation and Classification}

\author{Abolfazl~Meyarian and Xiaohui~Yuan \\
Department of Computer Science and Engineering\\
University of North Texas, Denton, TX, USA 76201\\
AbolfazlMeyarian@my.unt.edu, xiaohui.yuan@unt.edu
}

\maketitle

\begin{abstract}
High-resolution aerial imagery allows fine details in the segmentation of farmlands. However, small objects and features introduce distortions to the delineation of object boundaries, and larger contextual views are needed to mitigate class confusion. In this work, we present an end-to-end trainable network for segmenting farmlands with contour levees from high-resolution aerial imagery. A fusion block is devised that includes multiple voting blocks to achieve image segmentation and classification. We integrate the fusion block with a backbone and produce both semantic predictions and segmentation slices. The segmentation slices are used to perform majority voting on the predictions. The network is trained to assign the most likely class label of a segment to its pixels, learning the concept of farmlands rather than analyzing constitutive pixels separately.  We evaluate our method using images from the National Agriculture Imagery Program. Our method achieved an average accuracy of 94.34\%. Compared to the state-of-the-art methods, the proposed method obtains an improvement of 6.96\% and 2.63\% in the F1 score on average.

\end{abstract}


\section{Introduction}

High-resolution remote sensing imagery enables improved quality of many applications such as land cover classification~\cite{liu2022raanet} and irrigation practice mapping~\cite{meyarian2022gradient}. The fine details of high-resolution images introduce new challenges~\cite{Yuan21,liang2021first}. 
Small objects and features that are invisible in low-resolution images become prominent in high-resolution images, which inversely impacts the correct delineation of object boundaries and classification. 
Moreover, spatial color and texture variations across regions within cropland, in these images become more noticeable and cause strong edges and features for a model to capture and process. Such information is not useful in all cases and may introduce confusion to the model learning process. 
Detection of the contour levees that depends on the differentiation of cropland borders from the levees is affected by such high resolution, as the homogeneous areas between the levees may look like individual crops. Therefore, the classifier needs to learn the cropland rather than focusing on localities. 

The use of contextual data has a strong correlation with the receptive field of a network. Using several convolutional layers of large kernel sizes, hence a deeper network increases the receptive field~\cite{selvaraju2017grad, liu2018receptive, badrinarayanan2017segnet, chen2018encoder,ronneberger2015u}. However, more layers to a network make training and optimization difficult, due to the gradient vanishing problem. Pooling operations also implicitly increase the receptive field but decrease the resolution, which causes difficulties in the recovery of the information. 
Alternatively, features in multiple scales have been used. Feature Pyramid network (FPN)~\cite{lin2017feature} uses extracted features at different scales using parallel streams of convolutional layers. DeepLabV3+~\cite{chen2018encoder} uses atrous convolution in large different rates to increase the size of the receptive field and capture long-distance sparse relations between pixels. 

\begin{figure}[!htb]
\centering
\small
\noindent\begin{tabular}{@{}cccc}
\includegraphics[height=1.5in]{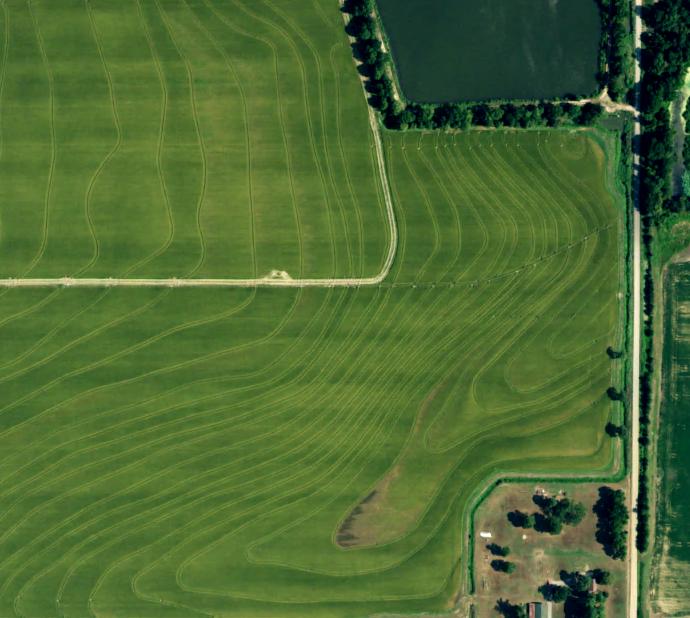}&
\includegraphics[height=1.5in]{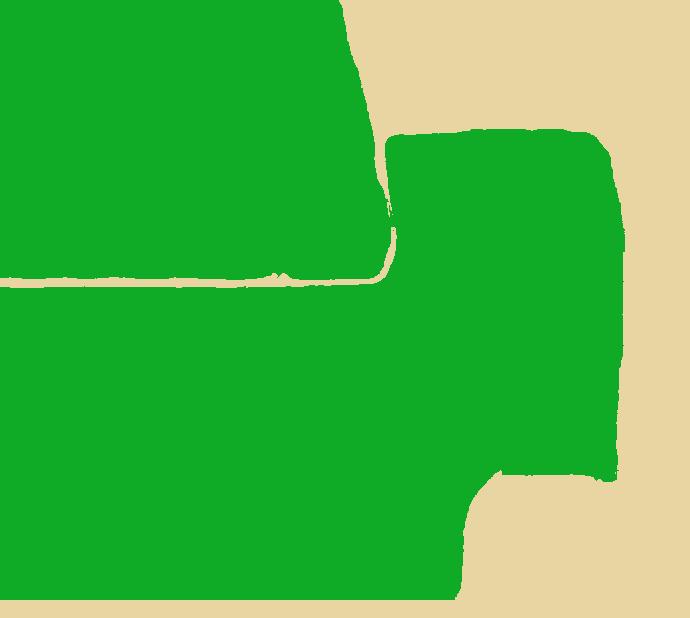}&
\includegraphics[height=1.5in]{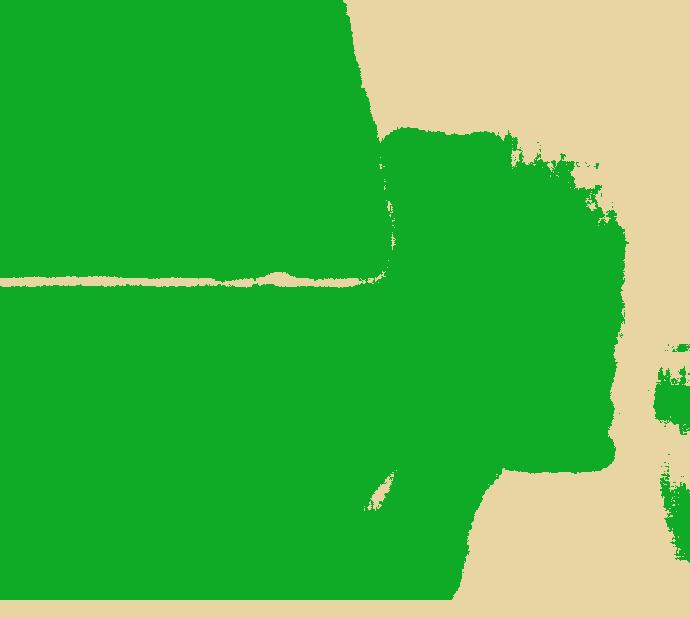}&
\includegraphics[height=1.5in]{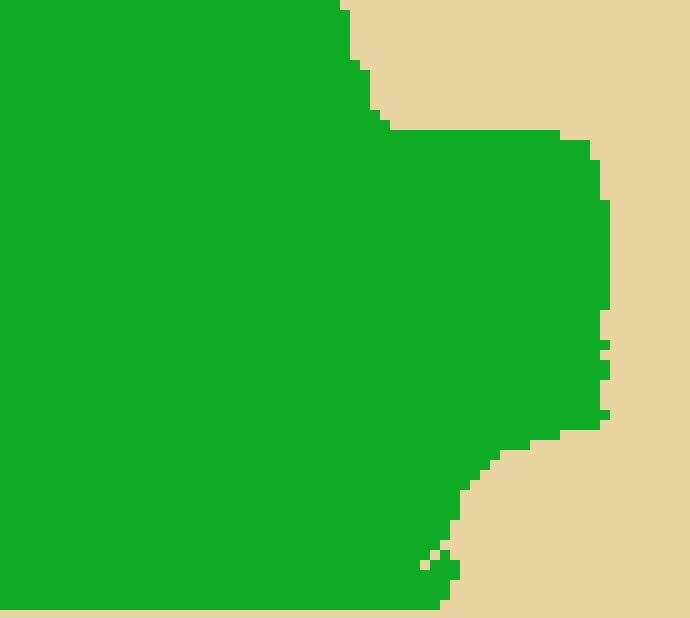} \\
\multicolumn{1}{c}{~(a) } & 
\multicolumn{1}{c}{~~(b) (96.75\%)} & 
\multicolumn{1}{c}{~~~(c) (94.55\%)} & 
\multicolumn{1}{c}{~~~(d) (94.97\%)}
\end{tabular}
\caption{{Farm land with contour levee and segmentation results. Farmlands with contour levee are depicted in green. (a) the aerial image, (b) the result by our proposed method, (c) the result by SegNet \cite{badrinarayanan2017segnet}, and (d) the result by Gradient CNN \cite{meyarian2022gradient}. The accuracy is reported in parentheses.}
\label{fig:3comp}}
\end{figure}

Using superpixel-based pre/post-processing is another common technique to improve prediction consistency. SAGNN~\cite{diao2022superpixel} uses SLIC segmentation to generate superpixels to build a graph, where features of the nodes are provided by a neural network. Predictions for each node are made through the convolutional head of the network. Gradient CNN~\cite{meyarian2022gradient} uses superpixels for majority voting. Applying superpixel segmentation on low-resolution images leads to improved computation cost at the expense of losing the quality of object boundaries (see Fig.~\ref{fig:3comp}(d)). This is partly caused by the unawareness of the semantic label of pixels. 

A trainable segmentation network overcomes the above problems and deep network-based segmentation methods have been developed~\cite{jampani2018superpixel, yang2020superpixel,xia2017w,ji2019invariant}. In addition to the color and position of the pixels, these methods use semantic labels to generate precise superpixel segmentation. Jampani et al.~\cite{jampani2018superpixel} created unified image segmentation and feature extraction method that uses a differentiable SLIC~\cite{achanta2012slic} to group pixels. Due to the iterative nature of SLIC, SSN has a high computational cost that requires the model to limit the similarity assessments of pixels to a small neighborhood. Therefore, SSN does not consider the long-distant relations between pixels in large objects. By using the reconstruction loss based on pixel attributes, these methods are more robust to the noises in the high-resolution image representations, which eliminates the need for downsampling. 
Xia et al.~\cite{xia2017w} developed a W-Net that uses two U-Nets for unsupervised image segmentation. The W-Net creates the image segments and reconstructs the image given the position and color features of the segments. It leverages the CRF method for detecting boundaries. 
 
This paper proposes an end-to-end trainable segmentation method that ensures the homogeneity of each segment and allows seamless integration with the downstream tasks. The method generates farmland segments and predicts their semantic labels. The training considers the prediction of pixels in the neighborhood. The trainable voting mechanism encourages the network to look at pixels in the context of the neighborhood. The inclusion of superpixel-in-loop helps refine the quality of the semantic areas to the closest agreement with the boundaries of the input image. Fig.~\ref{fig:3comp}(b) illustrates an example of the prediction result and our proposed method achieved a much-improved accuracy.

The rest of this paper is organized as follows: Section~\ref{sec:method} presents the details of our proposed method. Section~\ref{sec:experimental} discusses the evaluation results, including performance analysis and a comparison study. Section~\ref{sec:conclusion} concludes this paper with a summary.

\section{Method}\label{sec:method}

\begin{figure}[!htb]
\centering
\small
\includegraphics[width=5.5in]{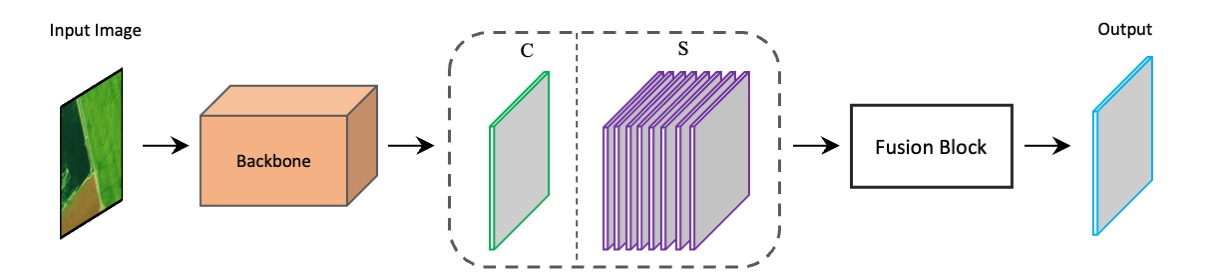} 
\caption{The architecture of VoteNet. The VoteNet uses a backbone to generate the segmentation slices $S$, and semantic prediction $C$, which contain $K$ and 2 slices, respectively. Then both $S$ and $C$ maps are passed to \textit{Fusion Block} for voting-based refinements, which generates an output with two slices.
\label{fig:net}}
\end{figure}

Our method extends DeepLabV3+~\cite{chen2018encoder}, shown as the backbone in Fig.\ref{fig:net}, by including a fusion block that has multiple parallel voting blocks. The network takes an RGB image as the input and generates semantic prediction output, denoted with $C$, and multiple slices, each of which contains a segment. These segments are denoted with $S$. The number of segment slices is $K$, which decides the maximum number of possible mutually exclusive fields in the image. The value of $K$ is fixed throughout the entire training and test phases.

In each slice of $S$, pixel values are in the range of $[0, 1]$, determining the probability of membership for each pixel to the segments. Therefore, for a pixel of the input image, there are $K$ probabilities provided in total. To obtain such probabilities, a softmax function is applied to the $K$ value at each coordinate of $S$ across all the slices. These segments of $S$ are used in the fusion block for majority voting. The semantic prediction of each segment is refined using the voting block. The parameter $K$ determines the maximum number of segments in the image. However, the network can identify fewer segments depending on the image at hand. 

If the maximum number of segments $K$ is large, the model tends to over-segment the image. When we have a small $K$, we face under-segmentation that is erroneous. The ideal value for $K$ is the number of fields in an image, which is unknown. Alternatively, we use the maximum number of fields in the training images as the value for $K$. In our experiments, ten is used, which gives satisfactory results. 

\begin{figure}[!htb]
\centering
\small
\includegraphics[width=3.5in]{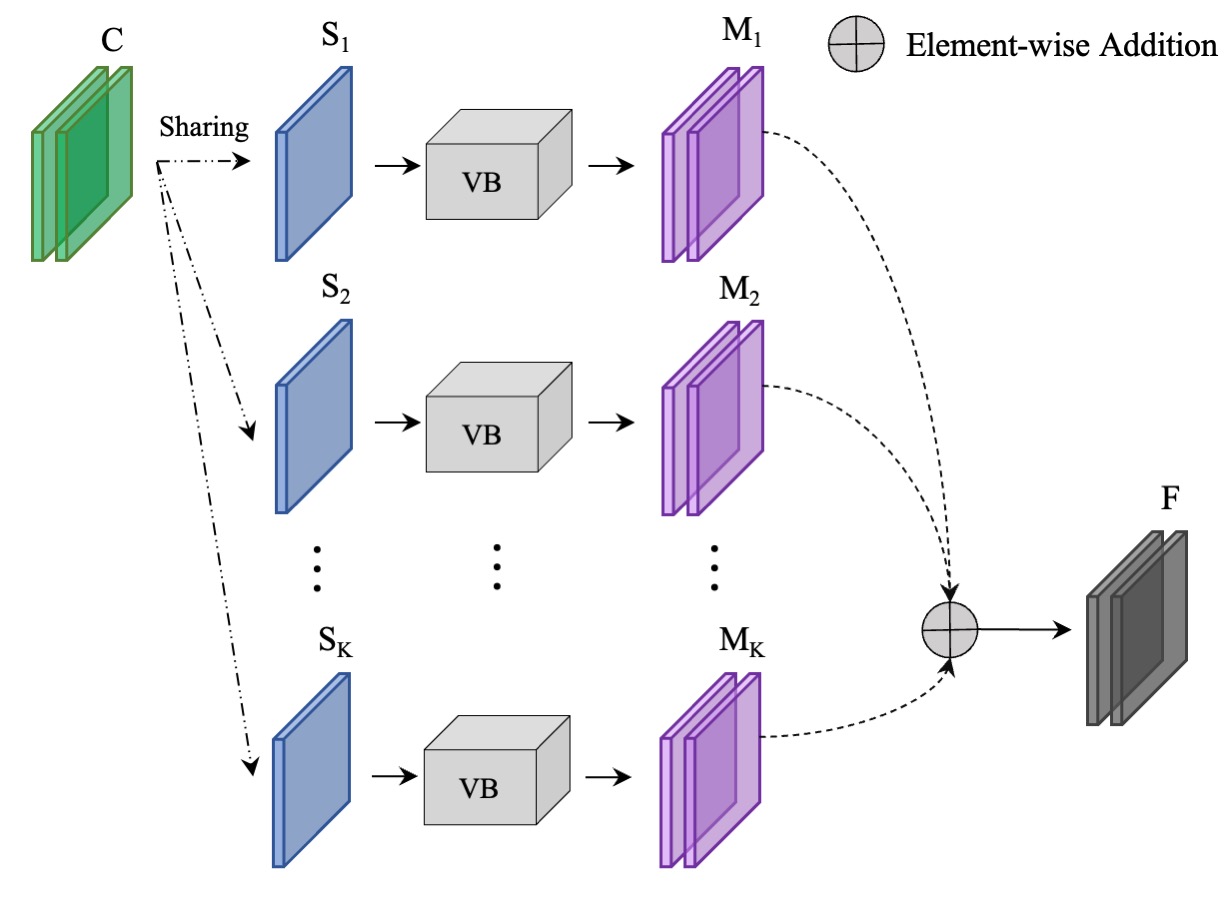}
\caption{Structure of the fusion Block.
\label{fig:fusionBlk}}
\end{figure}

We devise a deep majority voting module using the segment slices S and semantic prediction C on the predictions using a fusion block. The structure of FB is illustrated in Fig.~\ref{fig:fusionBlk}. The FB creates a voting process for each segment slice using a Voting Block (VB). Given an input segment $S_i$ and the segmentation prediction $C$, the voting block generates a refined segmentation $M_i$. The FB collects the refined features and performs an element-wise summation to obtain the refined segmentation map $F$ for the image. 

\begin{figure}[!htb]
\centering
\small
\includegraphics[width=4in]{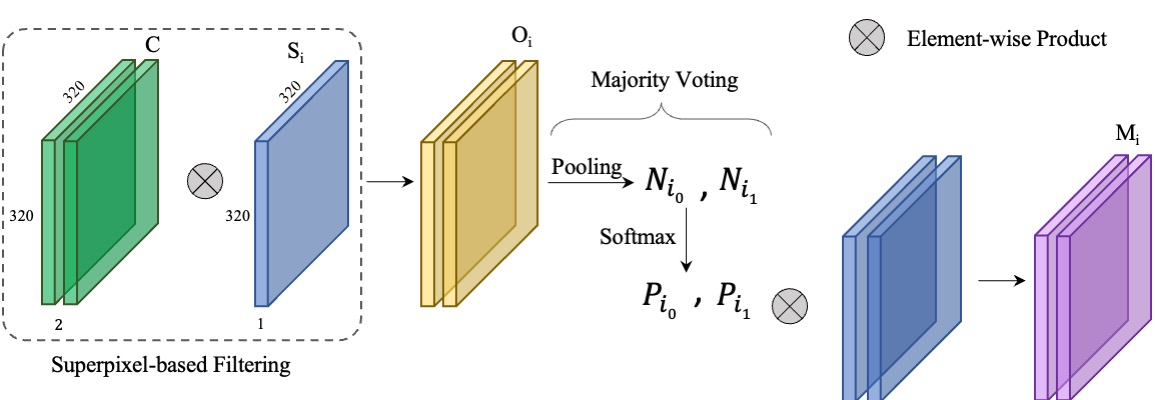}
\caption{Structure of the voting Block.
\label{fig:VotingBlk}}
\end{figure}

Each voting block inside the FB computes the number of labels and assigns the most frequent one to the entire segment, i.e., majority voting. Fig.~\ref{fig:VotingBlk} depicts its structure. To achieve majority voting, it first extracts the area of interest for $S_i$, through element-wise multiplication of $S_i$ and $C$, denoted as $O_i$. A summation of pixels across the width and height dimensions is performed on the product $O_i$ to get the count of pixels in segment $S_i$ belonging to the background and contour classes, denoted by $N_{i_b}$ and $N_{i_c}$, respectively. The class probability is computed by applying softmax to $N_{i_b}$ and $N_{i_c}$, which are broadcasted to $S_i$ through an element-wise multiplication of $S_i$ and each class probability. The element-wise products are concatenated and form the semantic segmentation mask for $S_i$. The voting block is applied to all segments in the fusion block. The segmentation mask for the image is obtained with an element-wise addition of all masks.

Our loss function consists of two parts: label-centric loss and region-centric loss. In the label-centric loss, we use weighted cross-entropy to compute the classification loss between the network output $F$ and ground-truth $L$ as follows: 
\begin{equation}
 -\eta L \log(\rho(F)) - (1-L)\log(1-\rho(F))  
\end{equation}
where $\eta$ is the weight given to the penalization of the network for misclassifying the contour class and $\rho(\cdot)$ is the sigmoid function. 

To optimize the initial segmentation map $C$ we compute an unweighted softmax cross-entropy of the segmentation map and the ground truth:  
\begin{equation}
\label{eq:softmax}
 - L \log(\rho'(C)) - (1-L)\log(1-\rho'(C))  
\end{equation}
where $\rho'$ is the softmax function. Given a feature map $A$, with $W\times H \times D$, where $D$ is the number of feature channels, we calculate the features of the segment centroid $A_k$ as follows:
\begin{equation}
 \label{eq:p}
     A_{k} = \frac{\sum_{i, j} S_{ijk}A_{ij}}{\sum_{i, j} S_{ijk}}
\end{equation}
The features of each pixel at $(i, j)$ are reconstructed using a weighted linear combination of centroid features~\cite{yang2020superpixel}:
\begin{equation}
    \hat{A}_{ij} = \sum_{k=1}^{K} S_{ijk}A_{k}.
\end{equation}
The contribution of each centroid is proportional to the membership of the pixel to that superpixel. The granularity error between the reconstructed and original features of pixels 
depends on the type of feature map considered for reconstruction. To reconstruct semantic labels $L$, we use softmax cross-entropy, similar to Eq.~\eqref{eq:softmax}. The label-centric loss term $\mathcal{L}_c$ is computed as follows:
\begin{eqnarray}
\label{eq:lc}
\mathcal{L}_c =  -L\log(\rho(F)^{\eta} \rho'(C)\rho'(\hat{L})) - (1-L) \log((1- \rho(F)) \\ \nonumber (1 - \rho'(C))(1- \rho'(\hat{L}))).
\end{eqnarray}

The region-centric loss term includes errors of partition, position, and color. The Partition Coefficient~\cite{bezdek1981cluster} determines the level of certainty of the membership of data records to their corresponding clusters. Given the membership probabilities $Q$ for the $i$th data sample provided by a clustering method, it is computed as follows:
\begin{equation}
\label{eq:PC}
    {PC} = \sum_{z=1}^{Z} Q_{iz}^2,
\end{equation}
where $Z$ is the number of clusters. The partition coefficient has a range of $(\frac{1}{Z}, 1)$. At the minimum score $\frac{1}{Z}$, we have the most uncertain clustering setting in which the data sample is equally probable to be part of any cluster. Therefore, maximizing this term is the goal to obtain the best clustering. 
In a segmentation result, each pixel belongs to only one image segment. However, if it is assigned to multiple segments, with almost equal affinity values, the optimization algorithm will issue gradients of the reconstruction losses for the contribution of the pixel to each segment it belongs to. This may lead to sub-optimal segmentation decisions. Partition Coefficient is used to force the network to assign each pixel to only one superpixel with high confidence and ensure a smooth optimization procedure and formation of image segments. The lost term is computed as follows:
\begin{equation}
\label{eq:PC}
\mathcal{L}_{PC} = \frac{1}{N} \sum_{i,j,k}^{H,W,K} S_{ijk}^2,
\end{equation}
where $N$ is the product of H, W, and K.

Similar to the reconstruction of the label, we use reconstruction loss on the position, $P$, and color, $\Psi$, features of the pixels to obtain high-quality image segments. To calculate the granularity error, we have the distance metric $D$ as follows:   
\begin{equation}
\label{eq:gran_p}
\frac{1}{M} \sum_{i,j}^{H,W}\| P_{ij} - \hat{P}_{ij} \|^2, ~~~
\frac{1}{M} \sum_{i,j}^{H,W} \| \Psi_{ij} - \hat{\Psi}_{ij} \|^2
\end{equation} 
where $M$ is the product of H and W, and $\hat{P}_{ij}$ and $\hat{\Psi}_{ij}$ are the reconstructed position and features of pixel at location $(i,j)$. The region-centric loss term $L_r$ is:
\begin{eqnarray}
\label{eq:lr}
\mathcal{L}_r = \frac{1}{M} \sum_{i,j}^{H,W}\| P_{ij} - \hat{P}_{ij} \|^2
+ \| \Psi_{ij} - \hat{\Psi}_{ij} \|^2 
+\frac{1}{N} \sum_{i,j,k}^{H,W,K} S_{ijk}^2
\end{eqnarray}
The loss function of our network is the weighted summation of the two terms as follows:
\begin{equation}
    \mathcal{L} =\lambda_c \mathcal{L}_c + \lambda_r \mathcal{L}_r,
\end{equation}
where $\lambda_c$ and $\lambda_r$ are weights.

\section{Experiments}\label{sec:experimental}

\subsection{Dataset and Settings}

Our dataset consists of images from the National Agriculture Imagery Program with a 1-meter resolution in three counties of Arkansas in 2015. The images cover an area of 400 million square meters. In our experiments, we augmented the dataset with rotations at $[5, 10, \ldots, 180]$ degrees to obtain more training data. 
\begin{figure}[!htb]
\centering
\small
\begin{tabular}{c}
\includegraphics[width=2in]{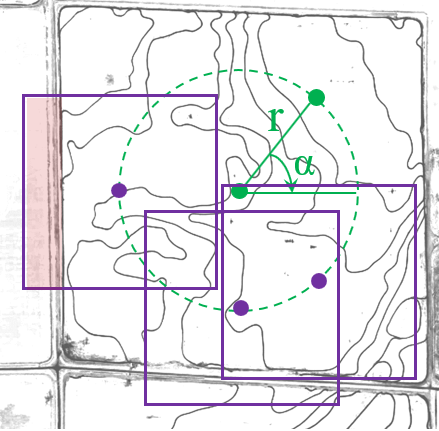} \\
\end{tabular}
\caption{The sampling technique for generating training image patches.
\label{fig:sampling}}
\end{figure}

To generate the training data, we compute the center of each farmland and take a neighborhood within a radius $r$. A point on the perimeter of the circle is selected in every $\alpha$ degree, which serves as the center of a sampled patch. The patches with at least 35\% of pixels marked as contour levees are kept. Fig.~\ref{fig:sampling} shows three patches outlined with squares. In our experiments, we used $0.1R$, $0.2R$, and $0.3R$ for r, where $R$ is the minimum of height or width of the farmland. We use a sliding window to generate patches of background pixels. As a result, over 178,000 patches are generated, from which 103,000 randomly selected patches are used for training.  

\begin{figure*}[!htb]
\centering
\small
\begin{tabular}{@{}cc}
\includegraphics[width=3.5in]{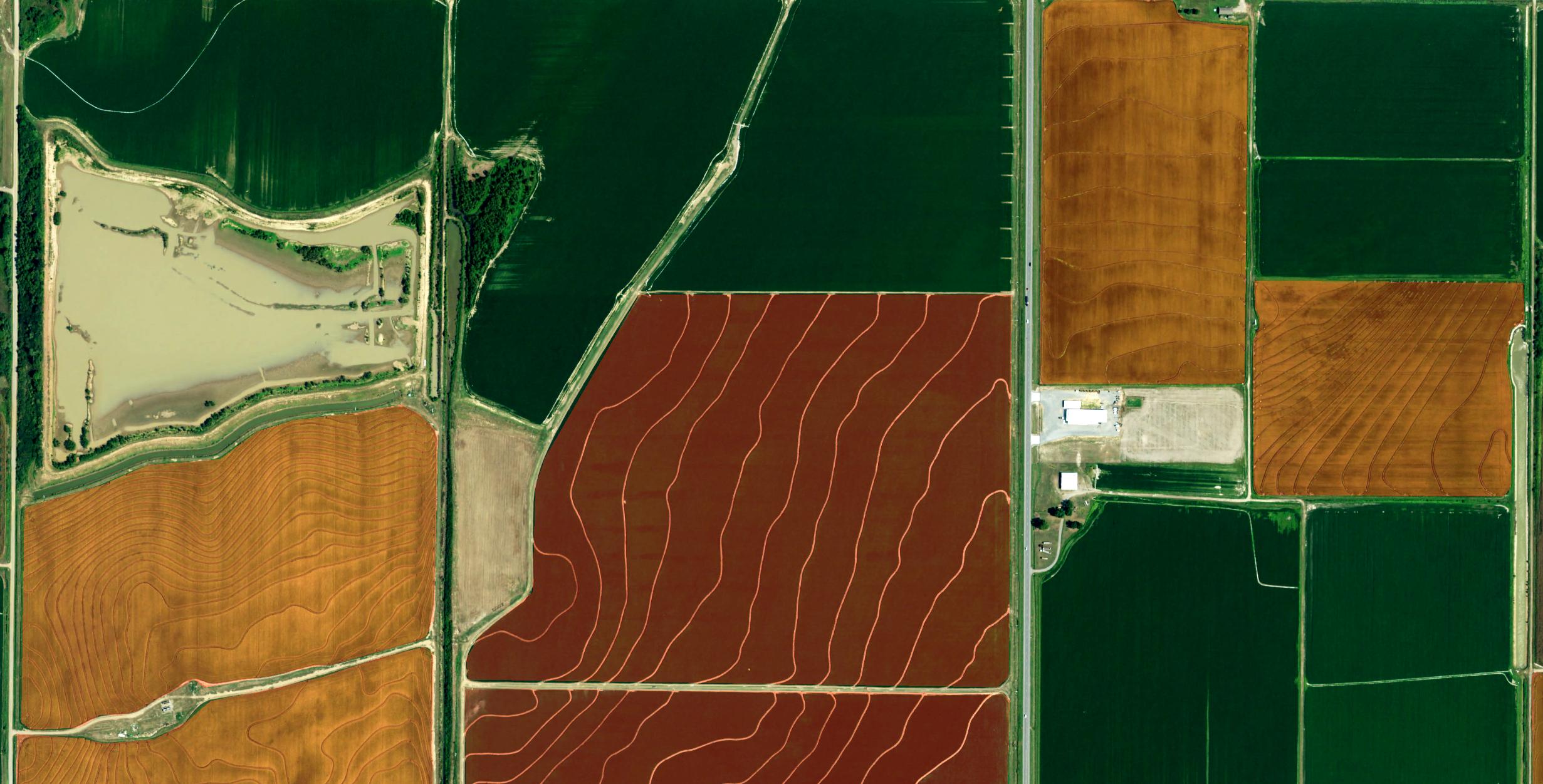} &
\includegraphics[width=3.5in]{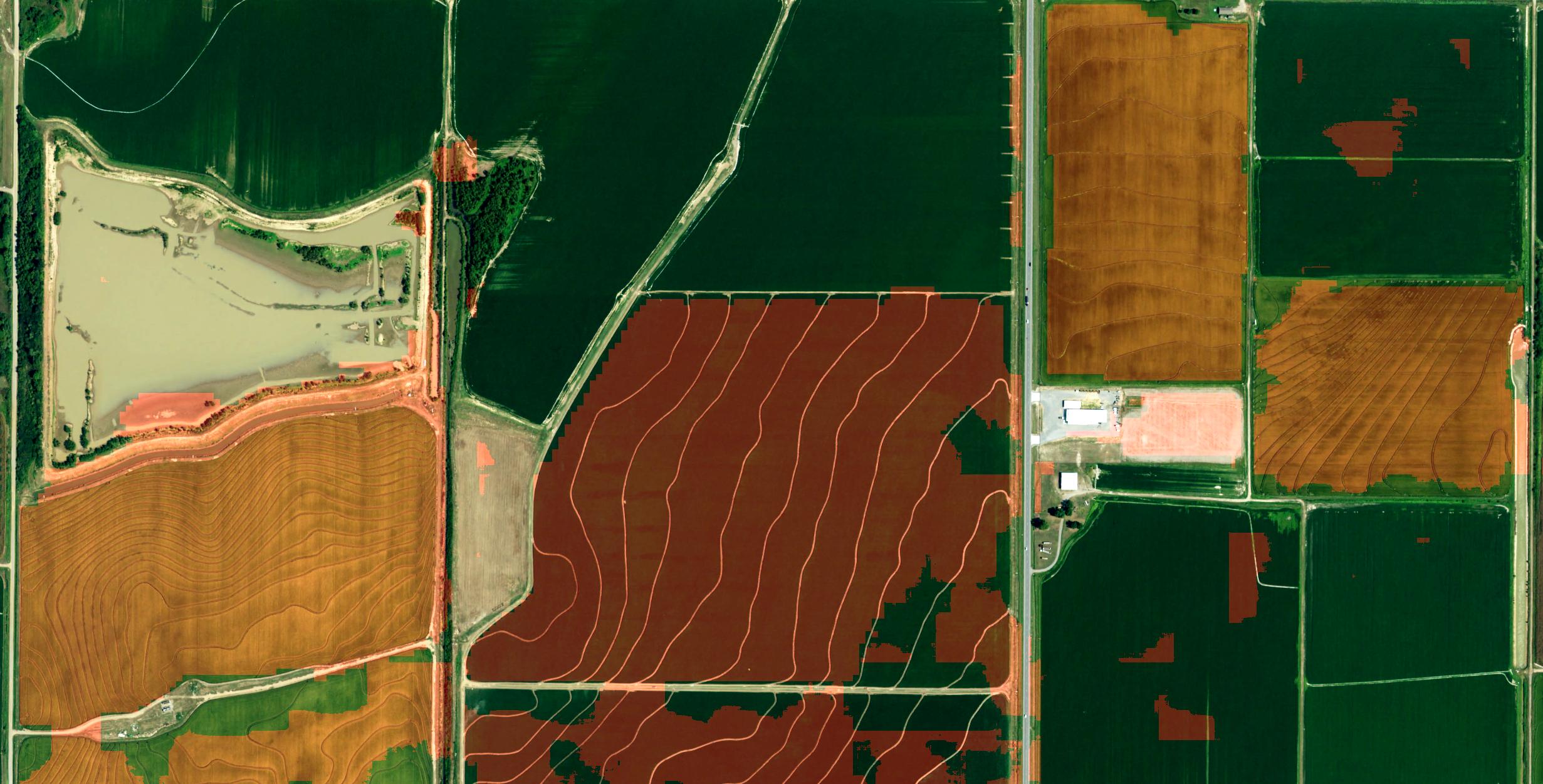} \\
(a) Ground-truth & (b) Gradient CNN ($92.7\%$)\\
\includegraphics[width=3.5in]{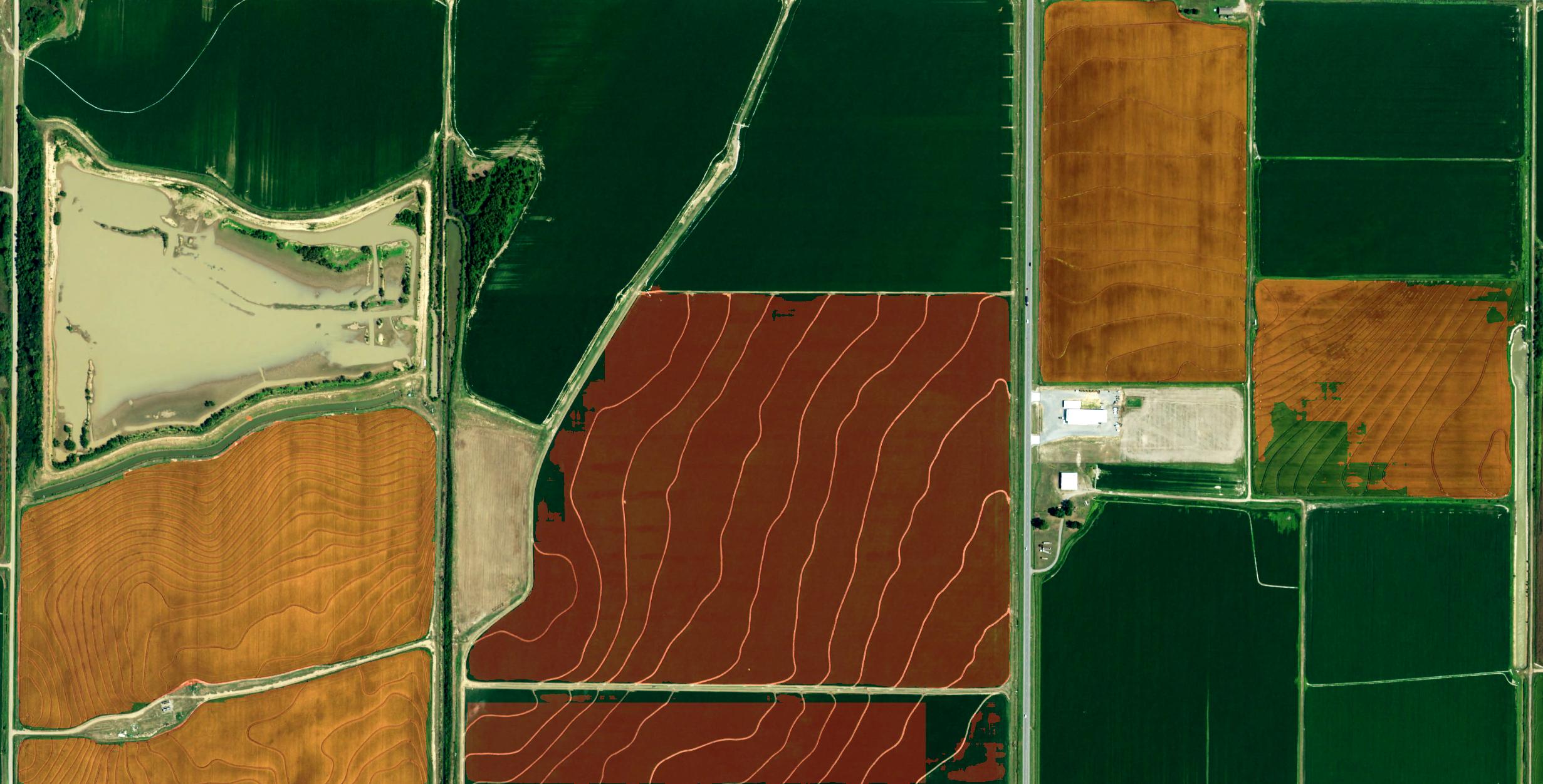} &
\includegraphics[width=3.5in]{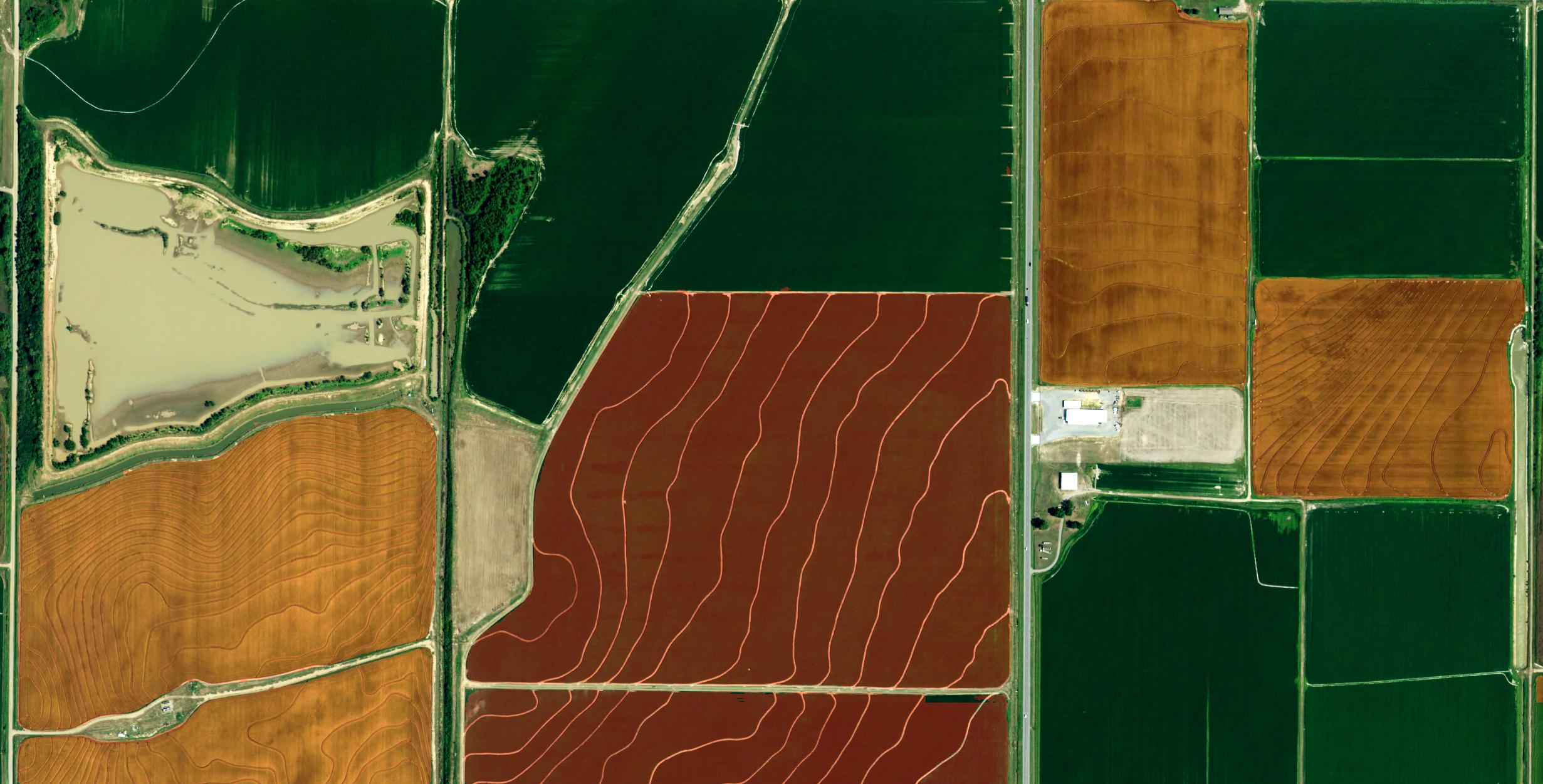} \\
(c) FPN-512 ($98.0\%$) & (d) VoteNet ($99.3\%$)
\end{tabular}
\caption{{An exemplar case and the results by Gradient CNN, FPN-512, and VoteNet(FPN). Regions in reddish color are farmlands with contour levee irrigation practices. The accuracy of each model is also reported in parentheses.}
\label{fig:5images}}
\end{figure*}

We use DeepLabV3+ as the backbone network. The optimization technique is the Adam with weights for computing moments at $0.9$ and $0.999$. The learning rate is initialized with $10^{-4}$ and changed to $10^{-5}$ after one epoch and remains the same for the rest of the training. Our model was trained with three epochs. 
In prediction, the connected components with at least 2,000 pixels are extracted for majority voting to avoid voting on small noisy segments. Our evaluation metrics include accuracy, F1-score, and Balanced Error Rate (BER, the average error of all classes).

\subsection{Evaluation of Window Stride}

The stride used to move the sliding window can generate overlapping or non-overlapping windows. Table~\ref{tab:stride} provides the average performance of the network concerning different strides $S$. The best results are highlighted with boldface font and the second best is underlined. The second column presents the number of patches with a size of $512\times512$. Given the size of the input image, having a stride of 512 is considered a non-overlapping scan.
\begin{table}[!htb]
\centering
\caption{Average performance using different stride sizes (S). The standard deviation is reported in parentheses.
\label{tab:stride}}
\begin{tabular}{c|c|ccc}
\hline
Stride                   & \#  &Accuracy  & BER &  F1 \\ 
\hline \hline
64    &5041   & \textbf{94.71} (0.03)	& \textbf{0.082} (0.002)	& \textbf{89.19} (0.12)\\ 
128   &1296   & \underline{94.70} (0.04) & \textbf{0.082} (0.002) & \underline{89.15} (0.12) \\ 
256   &324    & 94.52 (0.12)	&0.082 (0.0) &89.05 (0.10)\\ 
512   &81     & 93.88 (0.11)	& 0.095 (0.003)	& 87.37 (0.31) \\ 
\hline
\end{tabular}
\end{table}

Among all the variations, the model achieved the best performance when the stride size was 64 pixels. The large overlap between sliding windows allows us to get more predictions to form a stronger decision. On the other hand, a small stride dramatically increases the processing time. And the difference in accuracy is less than 0.3\% between strides 64, 128, and 256. Compared to the stride at 128, the time spent on testing increased by about four times at the stride size of 64. To balance the computational cost and performance, we use a stride of 256 in the rest of our experiments.

\subsection{Analysis of Loss Terms}

To evaluate the contribution of label- and region-centric loss terms, we normalize the range of all loss terms to $[0-1]$. In the baseline case, one is used for $\lambda_c$ and $\lambda_r$, i.e., an equilibrium between the label and region-centric components. For each term, we increased or decreased its weight to 1.6 and 0.4, while keeping the rest unchanged. Reducing $\lambda_c$ or increasing $\lambda_r$ is equivalent to paying more attention to region-centric losses. This case makes the model more similar to traditional techniques of segmentation which heavily rely on the color and position of the pixels. A decrement of $\lambda_r$ or increment of $\lambda_c$ creates more of a label-guided image segmentation, ignoring the color and position components. We have also added the dynamic range of the total losses, referred to as Loss Val. Again, the best value of each column is shown with boldface font, and the second best is underlined.

\begin{table}[!htb]
\centering
\caption{Performance comparison of using different loss weights.
\label{tab:settings}}
\begin{tabular}{cc|c|ccc}
\hline
\multicolumn{1}{c|}{Param.} & Val. & Loss Val. & \multicolumn{1}{c}{Accuracy}  & \multicolumn{1}{c}{BER} & F1 \\ 
\hline \hline
\multicolumn{2}{l|}{Baseline} & 4   & \underline{93.83} (0.06) & 0.088 (0.004)	& \underline{87.33} (0.42) \\ 
\hline
\multicolumn{1}{c|}{\multirow{2}{*}{$\lambda_c$}}  & 0.4 & 2.8 & 90.28 (0.26)	& 0.110 (0.002)	& 82.59 (0.41)  \\

\multicolumn{1}{c|}{}     & 1.6 & 5.2   & 93.75 (0.50)	& \underline{0.087} (0.003)	& 87.29 (1.05)  \\ 
\hline
\multicolumn{1}{c|}{\multirow{2}{*}{$\lambda_r$}}  &0.4 & 2.8 &\textbf{94.34} (0.23)	& \textbf{0.083} (0.001)	 & \textbf{88.38} (0.51) \\

\multicolumn{1}{c|}{}     & 1.6 & 5.2   & 92.23 (0.48)	& 0.099 (0.006)	& 85.21 (0.92)  \\   
\hline
\end{tabular}%
\end{table}

Table~\ref{tab:settings} reports the average performance of the trained models using different weight combinations. A decrease of $\lambda_c$ or an increase of $\lambda_r$ makes the model more prone to noises in color, therefore, a loss of performance in comparison to the baseline. In the case of decreasing the $\lambda_r$, a performance boost is observed. This shows that the use of semantic labels alongside the color and position of pixels leads to higher precision in segmentation. Worth noting that in this configuration the model experiences a lower dynamic range for the total loss function compared to the baseline. Theoretically, it was expected that due to the higher contribution of label-centric loss an improvement occurs, however, a larger value for the total loss did not allow the network to obtain such improvement. Such instability of the network in this case compared to the baseline is also confirmed by the larger values of the standard deviation.

\subsection{Performance Comparison}

Table~\ref{tb:county} reports the performance comparison of the VoteNet with two different backbones against the state-of-the-art methods. Each method was trained three times and the average performance is reported. Each method is evaluated in terms of accuracy, BER, and F1 score. Different from accuracy and F1, a smaller BER indicates better performance.

\begin{table}[!htb]
\centering
\caption{Performance comparison of VoteNet with state-of-the-art on 18 tiles of Arkansas, Woodruff, and Lonoke counties. The methods identified with \textbf{$^\star$} are trained with images of size $300\times300$. 
\label{tb:county}}
\begin{tabular}{l|ccc}
\hline
\multicolumn{1}{c|}{Method}    & Accuracy  & BER  & F1 \\ \hline \hline
U-Net~\cite{ronneberger2015u} & 83.59 (0.30)	& 0.185 (0.004)	& 70.29 (0.49) \\
SegNet~\cite{badrinarayanan2017segnet} & 84.26 (1.17) & 0.196 (0.002) & 69.30 (0.003) \\
Gradient CNN$^\star$~\cite{meyarian2022gradient}   & 89.88 (0.18)	& 0.136 (0.005) & 80.62 (0.20)\\
FPN~\cite{seferbekov2018feature} & 91.24 (1.74)	& 0.145 (0.033)	& 81.15 (4.40)	\\
DeepLabV3+~\cite{chen2018encoder} &\underline{93.56} (0.46) & \underline{0.110} (0.010) &\underline{86.11} (0.53) \\
{\bf VoteNet} (FPN)  & 93.30 (0.13)	& 0.094 (0.001)	& 86.80 (0.22)\\
{\bf VoteNet} (DeepLabV3+) & \textbf{94.34} (0.23)	& \textbf{0.083} (0.001)	 & \textbf{88.38} (0.51) \\
\hline 
\end{tabular}
\end{table}

Among the compared methods, VoteNet obtained the best performance. Compared to the second-best (highlighted with an underscore), VoteNet improves the accuracy, BER, and F1 by 0.83\%, 24.54\%, and 2.63\%, respectively. 
When FPN is used as the backbone network of our VoteNet, we observe a 2.2\% improvement of accuracy. Similarly, DeepLabV3+-based VoteNet achieves an improvement of 0.83\% in contrast to DeepLabV3+. It is evident that our VoteNet framework improves the segmentation performance.
The number in parenthesis reports the standard deviation (STD). Both versions of VoteNet have a small STD. This implies great consistency in the segmentation results. Gradient CNN also demonstrates consistency, which is a result of majority voting-based postprocessing.  

Fig.~\ref{fig:5images} illustrates the results of some representative methods. Although the FPN demonstrates a higher level of consistency and accuracy in the predictions compared to Gradient CNN, however, it still lacks quality in the prediction of cropland boundaries. Such cases are shown in Fig.~\ref{fig:5images}(c) in the middle-left and middle-bottom crops. On the other hand, VoteNet(FPN) using the same backbone, presents a high-quality boundary prediction with an accuracy of $99.3\%$ that shows its improved understanding of the concept of croplands as individual objects.

\subsection{Training Time Efficiency}

For all methods except the VoteNet, we use the predefined softmax cross-entropy loss. All methods are trained and evaluated on our machines which use the Ubuntu 20.04 operating system, an Intel i7-10700k CPU, and 32GB of memory, equipped with an Nvidia RTX-3090 with 24GB memory. Our method is implemented using Python 3.8 and Tensorflow 2.4.  

Table~\ref{tb:time} reports the Training Time Per Epoch (TTPE) and the total number of trainable parameters of our method and the state-of-the-art methods. The methods are ordered according to the TTPE. The training time of VoteNet is mainly affected by the efficiency of the backbone used, while the number of parameters depends only on the number of slices. For the case of using FPN as the backbone, the overhead training time of VoteNet is about 1 hour, compared to the training time of the FPN. The time complexity is comparable. The number of additional parameters is 0.266 million. 

\begin{table}[!htb]
\centering
\caption{Performance comparison of VoteNet with state-of-the-art models in terms of Training Time Per Epoch~(TTPE) and the number of parameters of the network. 
\label{tb:time}}
\begin{tabular}{l|cc}
\hline
\multicolumn{1}{c|}{Method}                   & TTPE (s)  & \# of Param. (M)\\ \hline \hline
Gradient CNN~\cite{meyarian2022gradient} & 3112 &10.710 \\
U-Net~\cite{ronneberger2015u} & 5991 & 31.031\\
{\bf VoteNet} (DeepLabV3+) & 8592  & 41.053\\
SegNet~\cite{badrinarayanan2017segnet} & 11407 &29.434\\
DeepLabV3+~\cite{chen2018encoder} & 14089 & 41.050 \\
FPN~\cite{seferbekov2018feature} &35376 	&29.700\\
{\bf VoteNet} (FPN)                 & 39950 & 29.705  \\ 
\hline
\end{tabular}
\end{table}

The VoteNet using DeepLabV3+ as the backbone used much less time than in training compared to DeepLabV3+. Compared to the FPN, DeepLabV3+ has a lower level of delay in the preparation of the final output, which uses a sequence of parallel streams of convolution. When VoteNet (DeepLabV3+) is trained with the custom training loops provided in Tensorflow, a 39\% improvement in TTPE is obtained while adding only 0.003 million parameters. VoteNet (DeepLabV3+) is among the top three most efficient methods.

\section{Conclusion}\label{sec:conclusion}

Detecting lands that use contour levees for irrigating rice crops in high-resolution remote sensing imagery is challenging. Farmlands represented in such a high level of detail create noise effects in the predictions, making it difficult to assess relations between pixels in long-range and accurately detect lands and their boundaries. 

This paper presents a voting network for farmland segmentation and classification. Our method generates high-resolution image segments from the color, position, and label of pixels used to perform majority voting. 
VoteNet obtains a performance gain of 9.63\% in the F1 score compared to the Gradient CNN. Moreover, using the edge-preserving voting mechanism improved the quality of the predictions on the field boundaries significantly. Compared to FPN and DeepLabV3+, VoteNet achieves a performance improvement of 6.96\% and 2.63\% in the F1 score, respectively, which demonstrates its potential of improving the performance of the existing state-of-the-art deep networks. According to our experiments, the addition of the VoteNet to a backbone does increase the number of parameters significantly, while it improves the segmentation accuracy. VoteNet also was ranked among the top-3 methods in terms of the TTPE showing its efficiency in terms of the training time complexity.  

Our experiments showed that using a smaller stride for prediction improves the overall accuracy of the VoteNet. However, the margin is not significant compared to the computational overhead. Analysis of the contribution of the loss terms also revealed that the label-centric term has a greater impact on the overall performance of VoteNet, compared to the region-centric term.

\bibliographystyle{unsrt}

\end{document}